\documentclass{article}

\usepackage{arxiv}

\usepackage[utf8]{inputenc} 
\usepackage[T1]{fontenc}    
\usepackage{hyperref}       
\usepackage{url}            
\usepackage{booktabs}       
\usepackage{amsfonts}       
\usepackage{nicefrac}       
\usepackage{microtype}      
\usepackage{lipsum}
\usepackage{graphicx}
\graphicspath{ {./images/} }

\title{Human-like visual computing advances explainability and few-shot learning in deep neural networks for complex physiological data}

\author{
 Alaa Alahmadi \\
  School of Computing\\
  Newcastle University \\
  Newcastle upon Tyne, United Kingdom \\
  \texttt{alaa.alahmadi@ncl.ac.uk} \\
   \And
 Mohamed Hasan \\
  School of Computing \\
  University of Leeds\\
  Leeds, United Kingdom \\
} 

\begin{document}

\maketitle

\begin{abstract}
Machine vision models, especially deep neural networks, are increasingly applied to physiological signal interpretation, including electrocardiography (ECG), yet they typically require large training datasets and provide limited insight into the causal features underlying their predictions. This lack of data efficiency and interpretability constrains their clinical reliability and alignment with human reasoning. Here, we demonstrate that a perception-informed pseudo-colouring technique, previously shown to enhance human interpretation of ECGs, can be leveraged to improve both explainability and few-shot learning in deep neural networks analysing complex physiological data.  We focus on a specific clinical use case, acquired drug-induced long QT syndrome (LQTS), which represents a challenging case study of heterogeneous and complex physiological data arising from diverse pharmacological mechanisms, where visual representations of LQTS are affected by varing heart rate and drug-induced ECG morphological changes.  Positive cases, where drug exposure precipitates life-threatening arrhythmias such as torsades de pointes, are inherently rare, resulting in limited labelled data. This setting provides a stringent testbed for evaluating the ability of machine learning models to generalise under extreme data scarcity and to learn clinically meaningful representations from small numbers of examples. By encoding clinically salient temporal features, such as QT-interval duration, into structured colour representations, the approach enables models to acquire discriminative and interpretable features from as few as one or five training examples. Using prototypical networks and a ResNet-18 architecture, we evaluate one-shot and few-shot learning performance on ECG image representations derived from single cardiac cycles and from full 10-second rhythms. Model explanations generated using local, model-agnostic interpretability methods reveal that pseudo-colouring guides human-like attention toward clinically meaningful ECG features while suppressing irrelevant signal components. We show that pseudo-colour-enhanced representations substantially improve classification accuracy, robustness, and interpretability under extreme data scarcity. Moreover, aggregating multiple cardiac cycles within a rhythm-level representation further enhances performance, mirroring human cognitive processes of perceptual averaging across heartbeats. Together, these findings suggest that incorporating human-like perceptual encoding into machine learning pipelines can bridge data efficiency, explainability, and causal reasoning in physiological signal analysis, with implications for broader applications in medical machine intelligence. 
\end{abstract}


\section{Introduction}

Statistical machine vision models have become increasingly effective in the analysis and interpretation of complex medical data, including imaging and signal-based modalities such as electrocardiograms (ECGs) and electroencephalograms (EEGs) \cite{frangi2023medical,strodthoff2020deep,craik2019deep}. These electro-physiological signals are among the most challenging data types for clinical interpretation, owing to their temporal complexity, inter-individual variability, and susceptibility to noise and artefacts \cite{benbadis2008errors,kashou2023ecg,macfarlane2017debatable}. Advances in artificial intelligence therefore hold significant promise for augmenting clinicians’ ability to visually monitor and interpret such signals, although several fundamental challenges remain unresolved \cite{topol2019deep}.

A central challenge in medical machine intelligence is achieving decision-making processes that are interpretable and trustworthy in a human-like manner—that is, intuitive, clinically grounded, and aligned with expert reasoning \cite{holzinger2017we,panayides2020ai,topol2019deep}. At the same time, many clinically important conditions are rare, heterogeneous, or sparsely labelled, limiting the availability of representative training data and undermining the reliability and generalisability of data-hungry deep learning models \cite{banik2021mitigating,topol2019deep}. These challenges are particularly pronounced for electro-physiological signals, which lack explicit spatial structure and exhibit substantial physiological overlap between normal and pathological patterns. As a result, clinically salient features are often entangled with benign signal variability, complicating abstraction, causal reasoning, and robust generalisation by machine vision systems \cite{kolk2023machine,hong2020opportunities}.

Here, we investigate a fundamentally different approach to representing and pre-processing ECG signals that draws inspiration from human perceptual strategies rather than purely statistical optimisation. Focusing on long QT syndrome (LQTS)—a clinically serious and visually subtle disorder associated with life-threatening arrhythmias such as torsades de pointes—we build on our prior work demonstrating that perception-informed pseudo-colouring can significantly enhance human ECG interpretation. We show that encoding clinically meaningful temporal features into structured colour representations enables deep neural networks to perceive ECG information in a more human-like manner. This approach simultaneously improves model accuracy, interpretability, and robustness, while enabling effective generalisation from very small numbers of training examples. Together, these results suggest that integrating human perceptual principles into machine learning pipelines offers a promising pathway toward more explainable, data-efficient, and clinically aligned machine intelligence for complex physiological signals

\section{Material and methods}

\subsection{Datasets}

We used the \textbf{ECGRDVQ database}, which contains 12-lead ECG signal recordings of 22 healthy subjects who participated in a randomized, double‐blind, 5‐period crossover clinical trial aimed at assessing the effect of four known QT-prolonging drugs versus placebo. The open ECG datasets are available online from the PhysioNet database \cite{goldberger2000physiobank}. 

The 10-second lead-II recording was selected from each 12-lead ECG, as this is typically used to measure the QT interval \cite{camm2008acquired}. The heart rates of the ECGs ranged from 40 to 96 beats per minute (bpm), and the QT-interval values ranged from 300 to 579 ms. We used the same ECGs (n = 5050) that were used in our previous studies evaluating the use of the pseudo-coloring technique when mapped according to the QT-nomogram (a clinical risk assessment method designed specifically for identifying patients at risk of drug-induced Torsades de Pointes (TdP) life-threatening arrhythmia according to heart rate \cite{chan2007drug}), where we evaluated the pseudo-coloring with human interpretation \cite{alahmadi2020pseudo}, and with a rule-based explainable algorithm \cite{alahmadi2021explainable}. 

As part of the clinical trial study methodology, medical experts have calculated the QT-interval and heart rate values for all ECGs. These QT/HR values were used as the ground truth for subsequently evaluating the model's binary classification performance. According to the QT/HR pair plots of all ECGs on the nomogram \cite{chan2007drug}, there were 180 positive ECGs showing a high risk of developing life-threatening TdP arrhythmia from drug-induced long QT syndrome, while the other negative ECGs (n = 4870) were below the nomogram line (no TdP risk). This significant imbalanced data is a common problem in medical machine learning research (where rare or positive cases typically have few examples/representations). Therefore, we specifically used few-shot and one-shot learning techniques (see section \ref{few-shot-method} for more details) to help overcome this problem using a new way of image representation and processing. 

\subsection{Image Representation \& Processing with Pseudo-coloring}

In this study, we re-visualize each ECG signal 10-second recording into four image representations, as follows:
\begin{itemize}
    \item a 256 x 256 single heartbeat image representation with and without pseudo-coloring.
    \item a 2048 x 256 10-second heart rhythm (i.e. multiple heartbeats) image representation, with and without using pseudo-coloring.
\end{itemize}

That is, we evaluated and compared the model performance across four image representations (Single Heartbeat vs. Heart Rhythm, with and without Pseudo-Color). More details on how to apply pseudo-coloring on ECG signals can be found here \cite{alahmadi2021explainable,alahmadi2020pseudo}. 

\subsection{Few-Shot and One-Shot Learning with Prototypical Networks}
\label{few-shot-method}





Few-shot learning is the challenge of training a model to make accurate predictions with only a small amount of data \cite{snell2017prototypical}. This new machine learning paradigm aims to bridge the gap between human common-sense few-shot learning and large-data machine learning \cite{snell2017prototypical}. We specifically employed this technique to evaluate the effectiveness of pseudo-color in producing accurate human-like predictions. In our previous work, humans need only a few examples of pseudo-colored ECGs per class (positive class = at risk of TdP, negative class = no TdP risk) during training time before the experiments to accurately classify the ECGs. That is, our aim here is to explore whether pseudo-coloring the ECGs may also endow the machine vision model with a similar 'human-like' process via few-hot and one-shot learning techniques. 

Most methods in few-shot learning use meta-learning algorithms, namely algorithms that learn to learn \cite{snell2017prototypical}. Here, we use prototypical networks (a groundbreaking approach that involves creating abstract representations, called prototypes, that capture the fundamental features of various classes in a dataset) as our main meta-learning algorithm, which has been shown to generalize more efficiently than other recent state-of-the-art meta-learning algorithms \cite{snell2017prototypical}. To build our model, we have utilized a pre-trained ResNet-18 Convolutional Neural Network (CNN). We used the EasyFSL open-source Few-Shot Learning library - licensed by MIT to design and run the experiments \cite{EasyFSL}.

Few-shot learning uses the N-Way K-Shot classification method, where N is the number of classes and K is the number of samples/shots from each class to train on. In this study, we used binary classification similar to our previous studies (i.e. ECGs were classified as 'At risk of Tdp' or 'No TdP risk'). As such, we use a 2-way 5-shot approach for the few-shot learning experiments and a 2-way 1-shot approach for the one-shot learning experiments.  

The concept of meta-learning involves sampling few-shot/one-shot tasks from a larger dataset, thus teaching the model algorithms to solve new, unseen tasks. We set the number of tasks for training the model to 500 to cover all data representations within the training dataset. We used 100 tasks to test and validate the model. The model was tested on 10 unseen query images in each task. Accuracy, sensitivity, specificity, precision, F1-score, and loss were calculated on each few-shot/one-shot task and then averaged across all tasks. 

\subsection{Experiments Design and Evaluation}

To better assess the model's accuracy and generalizability, we ran the few-shot and one-shot learning experiments using a 5-fold cross-validation method. Evaluation metrics were calculated for each fold and then averaged across the five folds. 

We conducted four experiments for few-shot learning and four experiments for one-shot learning to compare the four image representations (i.e., a single heartbeat vs. a 10-second heart rhythm with and without pseudo-coloring). 

\subsection{Explainability of Model Predictions}



We use the Local Interpretable Model-Agnostic Explanations (LIME) method to help explain the ECG features used in the model predictions and assess the model's general robustness and causality by applying it to each ECG case across the four different image representations.

\section{Results}


The study's results indicate that using pseudo-coloring to pre-process ECG images can significantly improve machine vision model accuracy when learning from as few as one example. This is achieved by representing the signal's continuous duration/amplitude using a sequence of colors, mapped according to clinical cut-off values. 
Pseudo-coloring ECGs help filter out irrelevant complex ECG features (e.g., distorted, variable T-wave morphologies) and help explain the machine vision causal reasoning behind the classification. 


The pseudo-coloring significantly helped increase the model sensitivity. When using one-shot learning, the sensitivity increased from 74.68\% to 83.14 when reading a single heartbeat image, and from 83.68\% to 95.54\% when reading a 10-second heart rhythm image. Similarly, with few-shot learning, the model sensitivity increased with pseudo-coloring from 82\% to 87.38\% with a single heartbeat image representation, and from 90.56\% to \%97.8 with a 10-second heart rhythm image representation.

Specificity was also increased with pseudo-coloring, whereas with one-shot learning, the effect was more significant, increasing from 86.86\% to 94.92\% when reading a single heartbeat image and from 90.96\% to 97.68\% when reading the 10-second heart rhythm image. With few-shot learning, the model specificity was also slightly increased with pseudo-coloring from 92.64\% to 96.92\% when reading a single heartbeat image, and from 96.82\% to 98.4\% when reading the 10-second heart rhythm image.


Overall, the experimental results, whether using pseudo-coloring or not, also show that reading more than one heartbeat within a 10-second heart rhythm image representation can significantly improve machine vision accuracy than reading a single heartbeat image. Using one-shot learning and pseudo-coloring, the accuracy significantly increased from 89.03\% for a single heartbeat image representation to 96.61\% for a 10-second heart rhythm image representation. Similar results were also found with few-shot learning, where the model accuracy also increased with pseudo-coloring from 92.15\% with a single heartbeat image representation to 98.1\% with a 10-second heart rhythm image representation.

Tables \ref{table1} and \ref{table2} show the study's experimental results using 5-fold cross-validation for the few-shot learning and one-shot learning, respectively. The evaluation metrics were calculated on each fold and then averaged across the five folds, tested on the same unseen ECG cases in each fold with and without pseudo-coloring for a proper comparison.

\begin{table}
 \caption{Few-shot learning experimental results using 5-fold cross-validation. Evaluation metrics below were averaged across the five folds.}
 \centering
  \begin{tabular}{lllll}
    \toprule
    & \multicolumn{2}{c}{Pseudo-Color}  & \multicolumn{2}{c}{No Color} \\
\cmidrule(r{4pt}){2-3} \cmidrule(l){4-5}
& Single Heartbeat & Heart Rhythm & Single Heartbeat &  Heart Rhythm \\
\midrule
Accuracy & 92.15 & 98.1 & 87.32 & 93.69 \\
Sensitivity & 87.38 & 97.8 & 82 & 90.56 \\
Specificity & 96.92 & 98.4 & 92.64 & 96.82 \\
Precision & 97.04 & 98.53 & 92.53 & 96.96 \\
F1-Score & 91.32 & 98.06 & 85.33 & 93.05 \\
Loss & 0.26 & 0.09 & 0.47 & 0.22 \\
\bottomrule
  \end{tabular}
  \label{tab:table}
\end{table}

\begin{table}
 \caption{One-shot learning experimental results using 5-fold cross-validation. Evaluation metrics below were averaged across the five folds.}
  \centering
  \begin{tabular}{lllll}
    \toprule
    & \multicolumn{2}{c}{Pseudo-Color}  & \multicolumn{2}{c}{No Color} \\
\cmidrule(r{4pt}){2-3} \cmidrule(l){4-5}
& Single Heartbeat & Heart Rhythm & Single Heartbeat &  Heart Rhythm \\
\midrule
Accuracy & 89.03 & 96.61 & 80.68 & 87.32 \\
Sensitivity & 83.14 & 95.54 & 74.68 & 83.68 \\
Specificity & 94.92& 97.68 & 86.68 & 90.96 \\
Precision & 95.70 & 97.94 & 86.84 & 91.65 \\
F1-Score & 87.23 & 96.40 &  77.82 & 86.27 \\
Loss & 0.45 & 0.11 & 0.66 & 0.47 \\
\bottomrule
  \end{tabular}
  \label{tab:table}
\end{table}

\section{Discussion}





\section{Discussion}

Two of the most persistent challenges facing artificial intelligence in medicine are the ability to interpret and understand model decisions in a manner that is intuitive and clinically meaningful, and the ability to generalise reliably from limited labelled data \cite{ghassemi2021false,holzinger2017we,panayides2020ai,topol2019deep}. Despite rapid advances in deep learning, current medical AI systems often remain brittle, data-hungry, and difficult to reconcile with human clinical reasoning. As a result, achieving clinically reliable, human-like AI remains an open challenge. Exploring alternative computational paradigms inspired by human perception and cognition may therefore be critical for improving the trustworthiness, interpretability, and real-world adoption of medical machine intelligence.

In this work, we focus on the interpretation of electrocardiogram (ECG) images, and in particular on acquired, drug-induced long QT syndrome (LQTS)—a clinically subtle but high-risk condition associated with torsades de pointes (TdP) and sudden cardiac death \cite{camm2008acquired}. This clinical setting presents a stringent test case for machine intelligence: positive cases are rare, heterogeneous, and often underrepresented in available datasets, while the relevant pathological features are temporally distributed and visually ambiguous within the raw signal. These characteristics make LQTS detection challenging not only for machines, but also for human experts, and therefore well suited for investigating human–machine alignment in perception and reasoning.

Our results demonstrate that modifying the representation of physiological signals—rather than solely increasing model complexity or dataset size—can substantially improve both learning efficiency and interpretability. By pseudo-colouring ECG signals to explicitly encode QT-interval duration, a clinically critical but visually subtle feature, we enable deep neural networks to acquire discriminative representations from as few as one or five labelled examples. Importantly, this pseudo-colouring strategy has previously been shown to significantly improve human ECG interpretation of LQTS \cite{alahmadi2020pseudo}, suggesting that the performance gains observed here arise from aligning machine perception with human perceptual strategies rather than introducing task-specific heuristics.

This shared representation appears to support a form of human-like causal reasoning in the models. Explainability analyses indicate that pseudo-colour-enhanced models attend preferentially to clinically meaningful regions of the ECG while suppressing irrelevant signal components, addressing a key limitation of many post hoc explainability methods that often highlight spurious correlations \cite{ghassemi2021false}. Rather than attempting to explain opaque internal representations after training, our approach embeds interpretability directly into the input space, allowing both humans and machines to reason over the same perceptually structured features. This distinction is particularly important for high-stakes clinical applications, where explainability must support meaningful clinical decision-making rather than merely satisfy technical criteria.

The observed improvement when using rhythm-level representations that aggregate multiple cardiac cycles further reinforces the parallels between human and machine perception. Clinicians routinely assess QT prolongation by visually averaging across beats to account for physiological variability, and we find that similar aggregation improves model robustness and accuracy under data scarcity. This convergence suggests that incorporating perceptual principles such as temporal integration and visual averaging may provide a powerful inductive bias for machine learning models operating on complex physiological signals.

While this study focuses on ECG interpretation and LQTS as a case study, the broader implications extend to machine intelligence for heterogeneous and sparsely labelled biomedical data. Many physiological and biomedical signals—ranging from EEGs to wearable sensor data—suffer from similar challenges of variability, limited positive cases, and weakly separable pathological features. Encoding domain-relevant temporal or amplitude-based abnormalities into perceptually meaningful visual structures may offer a general strategy for improving few-shot learning, robustness, and interpretability across medical domains. As such, apart from improving model accuracy, this study demonstrates how  using a science-of-perception-based approach can help develop human-like abstract perception and reasoning within machine vision models, supporting the model to generalize using only a small amount of training data. Our approach draws from human perception theories, particularly pre-attentive processing, to simplify visual complexity with pre-attentive properties (e.g., color, form, and spatial positioning) that cannot be decomposed into simpler features \cite{gilchrist2009psychology}. 

\begin{figure}
\centering
\includegraphics[width=\textwidth]{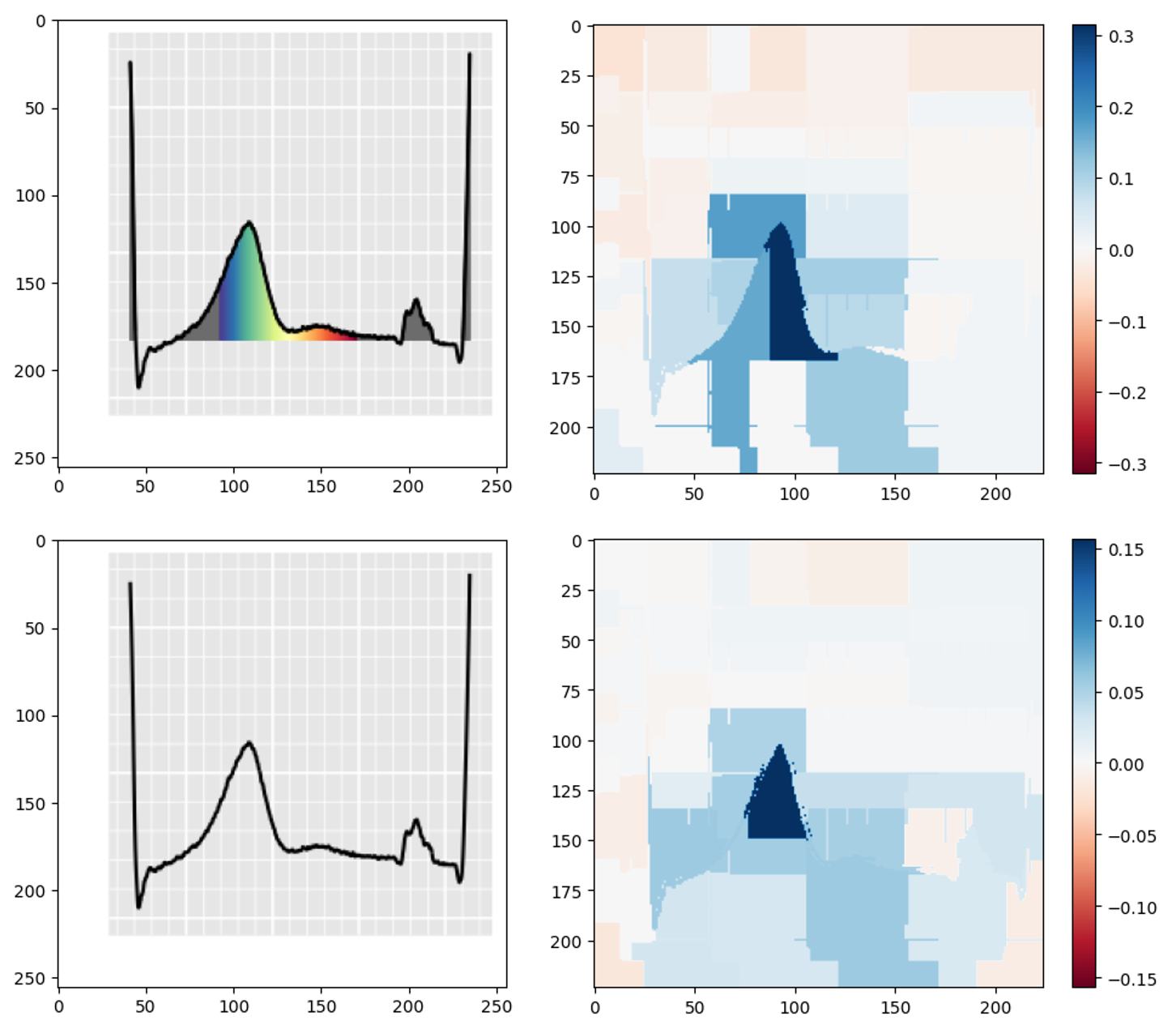}
\caption{Few-shot learning of single heartbeat showing normal QT-interval with no risk of TdP. With pseudo-coloring, the model focuses on both the T-wave morphology and duration (represented by the color), with more emphasis on the T-peak-T-end duration (known to be a significant clinical feature in assessing the risk of LQTS \cite{bhuiyan2015t,morin2012relationships}). Without pseudo-color, the model focuses mainly on the T-wave morphology.}
\label{fig1}
\end{figure}

\begin{figure}
\centering
\includegraphics[width=\textwidth]{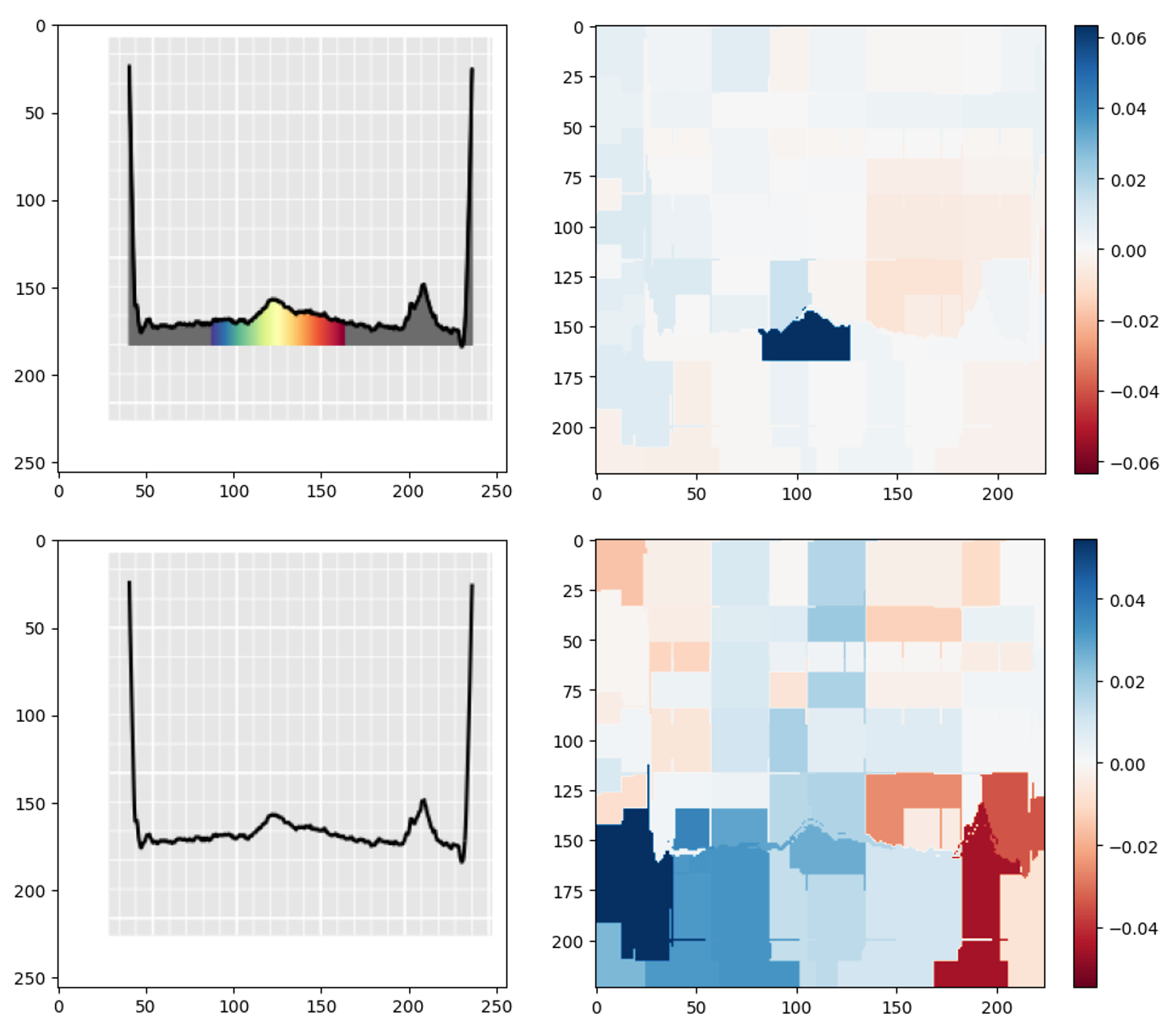}
\caption{Few-shot learning of single heartbeat showing an abnormal QT-interval prolongation at risk of TdP. With pseudo-coloring, the model focuses mostly on the T-wave area (which represents the T-wave morphology as well as the prolonged QT duration shown by the yellow-orange-red color gradient). Whereas without pseudo-color, the model incorrectly focuses on other irrelevant ECG features, in addition to the T-wave morphology.}
\label{fig2}
\end{figure}

\begin{figure}
\centering
\includegraphics[width=\textwidth]{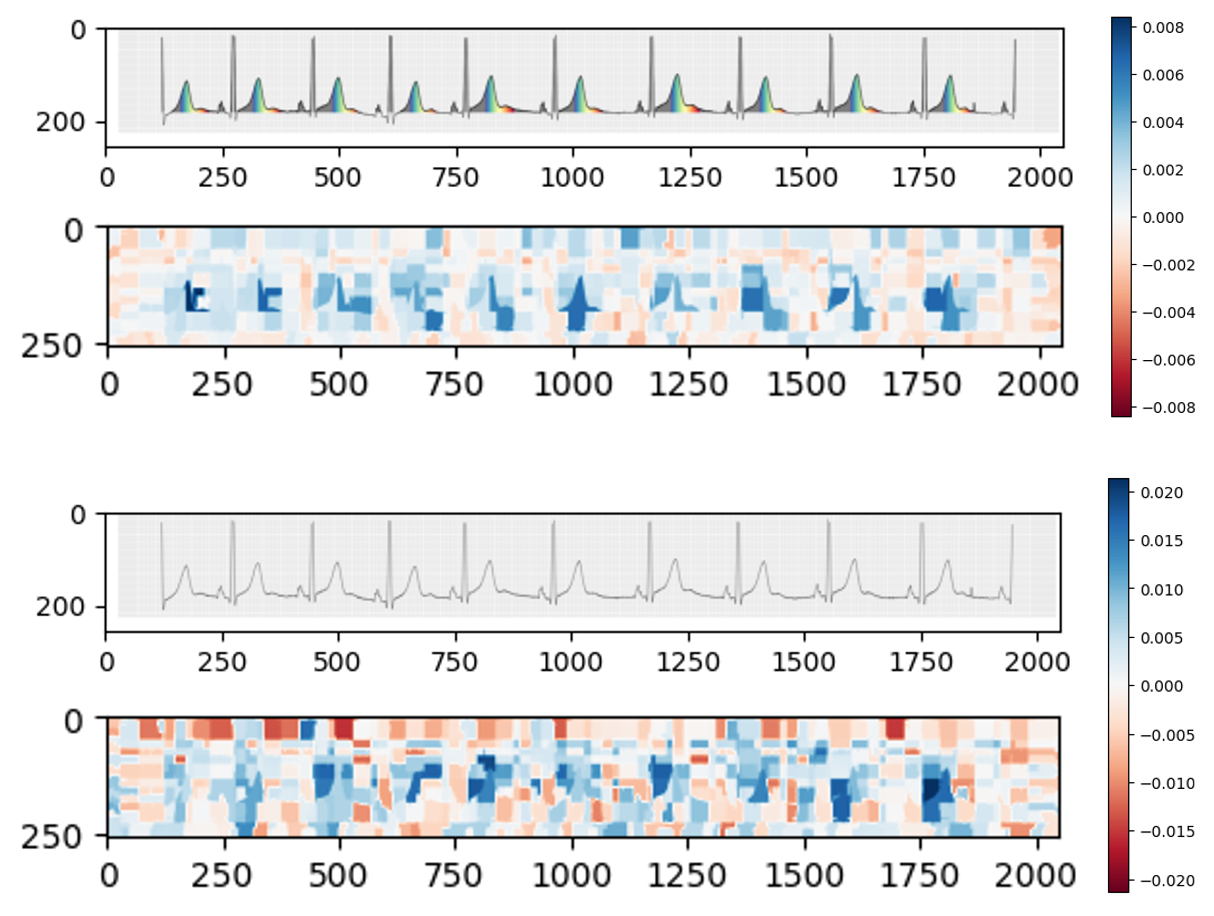}
\caption{Few-shot learning of a 10-second heart rhythm image representation showing normal QT-interval with no TdP risk. Through pseudo-coloring, the model extracted the most clinically significant features (T-wave morphology and duration) and filtered out irrelevant signal features. Without color, the model extracted several clinically irrelevant features - hindering machine abstract and causal reasoning.}
\label{fig3}
\end{figure}

\begin{figure}
\centering
\includegraphics[width=\textwidth]{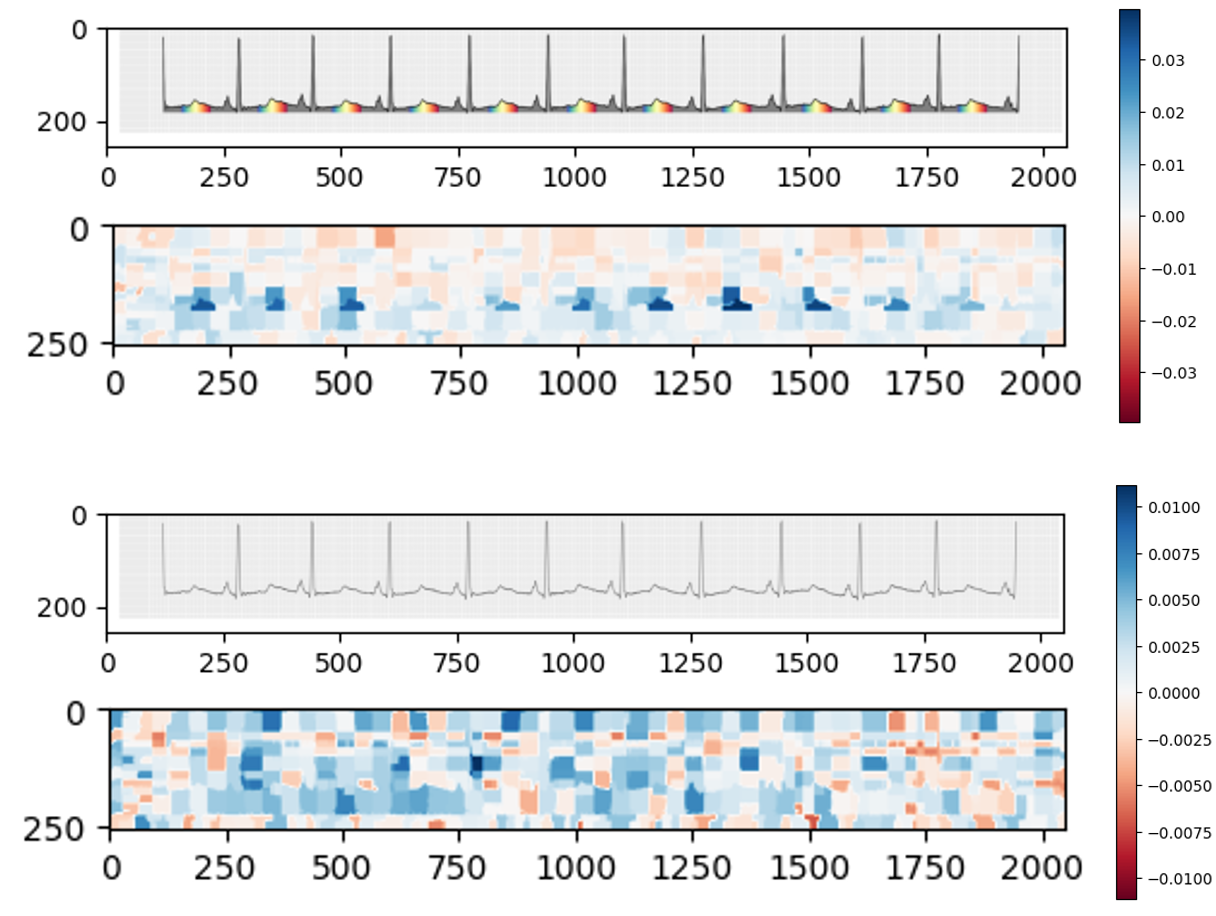}
\caption{Few-shot learning of a 10-second heart rhythm image representation showing abnormal QT-interval at risk of TdP. Using pseudo-coloring, the model extracted the most clinically significant features (T-wave morphology and duration) and filtered out irrelevant signal features. Without color, the model extracted several clinically irrelevant features - hindering machine abstract and causal reasoning.}
\label{fig4}
\end{figure}

Furthermore, the study shows that pseudo-coloring can help the machine vision feature extraction process by filtering out irrelevant complex ECG features (e.g., distorted, variable T-wave morphologies commonly associated with LQTS) and helping explain machine vision causal reasoning behind the classification. For example, using the explainable LIME method, Figure \ref{fig1} shows an example of a few-shot task analyzing a single heartbeat image showing normal QT-interval with no TdP risk; with pseudo-coloring, the model focuses on both the T-wave morphology and duration (represented by the color), with more emphasis on the T-peak-T-end duration (known to be a significant clinical feature in assessing the risk of LQTS \cite{bhuiyan2015t,morin2012relationships}); without pseudo-color, the model focuses mainly on the T-wave morphology. On the other hand, extracting abnormal LQTS features at risk of TdP from a single heartbeat image becomes significantly more challenging without the use of pseudo-coloring (as shown in Figure \ref{fig2}). The study also showed that analyzing more than one heartbeat within a 10-second heart rhythm image representation can significantly improve machine vision accuracy than analyzing a single heartbeat image - a process that we have previously shown to also help human interpretation in detecting long QT syndrome by visually averaging the QT-interval length across multiple heartbeats \cite{alahmadi2019can}. With few-shot learning, the pseudo-coloring helps the model extract the most clinically significant features associated with the condition (T-wave morphology and duration), filtering out irrelevant complex signal features that commonly hinder ECG interpretation (as shown in Figure \ref{fig3} representing a normal QT case, and in Figure \ref{fig4} showing abnormal QT case at risk of TdP).

\begin{figure}
\centering
\includegraphics[width=\textwidth]{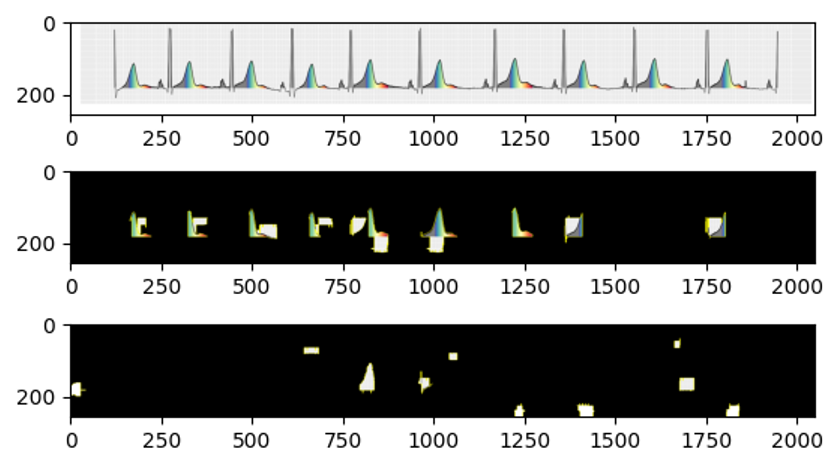}
\caption{One-shot learning of a 10-second heart rhythm image representation showing normal QT-interval with no TdP risk. Through pseudo-coloring, the model extracted the most clinically significant features (T-wave morphology and pseudo-colored duration) and filtered out clinically irrelevant signal features. Without pseudo-color, the model completely failed to extract any features that were clinically significant.}
\label{fig5}
\end{figure}

\begin{figure}
\centering
\includegraphics[width=\textwidth]{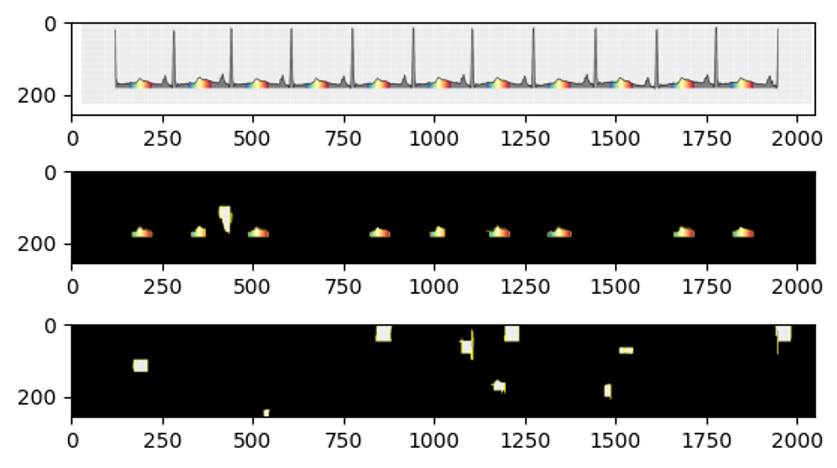}
\caption{One-shot learning of a 10-second heart rhythm image representation showing abnormal QT-interval at risk of TdP. Through pseudo-coloring, the model extracted the most clinically significant features (T-wave morphology and pseudo-colored duration) and filtered out clinically irrelevant signal features. Without pseudo-color, the model completely failed to extract any features that were clinically significant.}
\label{fig6}
\end{figure}

The effectiveness of pseudo-coloring was more significant with one-shot learning experiments. The model was able to extract the clinical features encoded in the pseudo-color learning from only one example, demonstrating generalizability. Without pseudo-color, however, the model completely failed to extract any clinically significant features (as illustrated by Figures \ref{fig5} and \ref{fig6} representing a normal QT case and abnormal QT case at risk of TdP, respectively).  

\section{Limitations and Future work}

Several limitations should be acknowledged. First, this work evaluates a specific pseudo-colouring strategy tailored to QT-interval abnormalities; future studies should investigate how different perceptual encodings generalise to other signal features and clinical tasks. Second, although few-shot performance is substantially improved, prospective validation on independent clinical datasets will be necessary to assess real-world deployment potential. Finally, while perception-informed representations reduce reliance on post hoc explanations, they do not eliminate the need for rigorous evaluation of model failure modes and biases.

\section{Conclusions}

In conclusion, our findings suggest that integrating human perceptual principles into machine learning pipelines can bridge key gaps between data efficiency, explainability, and clinical relevance. By enabling machines to learn and reason from representations that mirror human perception, this work points toward a complementary pathway to human-like machine intelligence—one grounded not only in statistical optimisation, but also in the science of perception and visual cognition.


\bibliographystyle{unsrt}  
\bibliography{references} 

@book{frangi2023medical,
  title={Medical Image Analysis},
  author={Frangi, Alejandro and Prince, Jerry and Sonka, Milan},
  year={2023},
  publisher={Academic Press}
}

@article{strodthoff2020deep,
  title={Deep learning for ECG analysis: Benchmarks and insights from PTB-XL},
  author={Strodthoff, Nils and Wagner, Patrick and Schaeffter, Tobias and Samek, Wojciech},
  journal={IEEE journal of biomedical and health informatics},
  volume={25},
  number={5},
  pages={1519--1528},
  year={2020},
  publisher={IEEE}
}

@article{craik2019deep,
  title={Deep learning for electroencephalogram (EEG) classification tasks: a review},
  author={Craik, Alexander and He, Yongtian and Contreras-Vidal, Jose L},
  journal={Journal of neural engineering},
  volume={16},
  number={3},
  pages={031001},
  year={2019},
  publisher={IOP Publishing}
}

@article{benbadis2008errors,
  title={Errors in EEG Interpretation and Misdiagnosis of EpilepsyWhich EEG Patterns Are Overread?},
  author={Benbadis, Selim R and Lin, Kaiwen},
  journal={European neurology},
  volume={59},
  number={5},
  pages={267--271},
  year={2008},
  publisher={S. Karger AG}
}

@article{kashou2023ecg,
  title={ECG interpretation proficiency of healthcare professionals},
  author={Kashou, Anthony H and Noseworthy, Peter A and Beckman, Thomas J and Anavekar, Nandan S and Cullen, Michael W and Angstman, Kurt B and Sandefur, Benjamin J and Shapiro, Brian P and Wiley, Brandon W and Kates, Andrew M and others},
  journal={Current problems in cardiology},
  pages={101924},
  year={2023},
  publisher={Elsevier}
}

@article{macfarlane2017debatable,
  title={Debatable issues in automated ECG reporting},
  author={Macfarlane, Peter W and Mason, Jay W and Kligfield, Paul and Sommargren, Claire E and Drew, Barbara and van Dam, Peter and Ab{\"a}cherli, Roger and Albert, David E and Hodges, Morrison},
  journal={Journal of electrocardiology},
  volume={50},
  number={6},
  pages={833--840},
  year={2017},
  publisher={Elsevier}
}

@book{topol2019deep,
  title={Deep medicine: how artificial intelligence can make healthcare human again},
  author={Topol, Eric},
  year={2019},
  publisher={Hachette UK}
}

@article{panayides2020ai,
  title={AI in medical imaging informatics: current challenges and future directions},
  author={Panayides, Andreas S and Amini, Amir and Filipovic, Nenad D and Sharma, Ashish and Tsaftaris, Sotirios A and Young, Alistair and Foran, David and Do, Nhan and Golemati, Spyretta and Kurc, Tahsin and others},
  journal={IEEE journal of biomedical and health informatics},
  volume={24},
  number={7},
  pages={1837--1857},
  year={2020},
  publisher={IEEE}
}

@article{holzinger2017we,
  title={What do we need to build explainable AI systems for the medical domain?},
  author={Holzinger, Andreas and Biemann, Chris and Pattichis, Constantinos S and Kell, Douglas B},
  journal={arXiv preprint arXiv:1712.09923},
  year={2017}
}

@incollection{banik2021mitigating,
  title={Mitigating data imbalance issues in medical image analysis},
  author={Banik, Debapriya and Bhattacharjee, Debotosh},
  booktitle={Data preprocessing, active learning, and cost perceptive approaches for resolving data imbalance},
  pages={66--89},
  year={2021},
  publisher={IGI Global}
}

@article{kolk2023machine,
  title={Machine learning of electrophysiological signals for the prediction of ventricular arrhythmias: systematic review and examination of heterogeneity between studies},
  author={Kolk, Maarten ZH and Deb, Brototo and Ruip{\'e}rez-Campillo, Samuel and Bhatia, Neil K and Clopton, Paul and Wilde, Arthur AM and Narayan, Sanjiv M and Knops, Reinoud E and Tjong, Fleur VY},
  journal={EBioMedicine},
  volume={89},
  year={2023},
  publisher={Elsevier}
}

@article{hong2020opportunities,
  title={Opportunities and challenges of deep learning methods for electrocardiogram data: A systematic review},
  author={Hong, Shenda and Zhou, Yuxi and Shang, Junyuan and Xiao, Cao and Sun, Jimeng},
  journal={Computers in biology and medicine},
  volume={122},
  pages={103801},
  year={2020},
  publisher={Elsevier}
}

@article{ghassemi2021false,
  title={The false hope of current approaches to explainable artificial intelligence in health care},
  author={Ghassemi, Marzyeh and Oakden-Rayner, Luke and Beam, Andrew L},
  journal={The Lancet Digital Health},
  volume={3},
  number={11},
  pages={e745--e750},
  year={2021},
  publisher={Elsevier}
}

@article{goldberger2000physiobank,
  title={PhysioBank, PhysioToolkit, and PhysioNet: components of a new research resource for complex physiologic signals},
  author={Goldberger, Ary L and Amaral, Luis AN and Glass, Leon and Hausdorff, Jeffrey M and Ivanov, Plamen Ch and Mark, Roger G and Mietus, Joseph E and Moody, George B and Peng, Chung-Kang and Stanley, H Eugene},
  journal={circulation},
  volume={101},
  number={23},
  pages={e215--e220},
  year={2000},
  publisher={Am Heart Assoc}
}

@article{camm2008acquired,
  title={Acquired long QT syndrome},
  author={Camm, A John and Malik, Marek and Yap, Yee Guan},
  year={2008},
  publisher={John Wiley \& Sons}
}

@article{chan2007drug,
  title={Drug-induced QT prolongation and torsades de pointes: evaluation of a QT nomogram},
  author={Chan, A and Isbister, GK and Kirkpatrick, CMJ and Dufful, SB},
  journal={QJM: An International Journal of Medicine},
  volume={100},
  number={10},
  pages={609--615},
  year={2007},
  publisher={Oxford University Press}
}

@article{alahmadi2021explainable,
  title={An explainable algorithm for detecting drug-induced QT-prolongation at risk of torsades de pointes (TdP) regardless of heart rate and T-wave morphology},
  author={Alahmadi, Alaa and Davies, Alan and Royle, Jennifer and Goodwin, Leanna and Cresswell, Katharine and Arain, Zahra and Vigo, Markel and Jay, Caroline},
  journal={Computers in Biology and Medicine},
  volume={131},
  pages={104281},
  year={2021},
  publisher={Elsevier}
}

@article{alahmadi2020pseudo,
  title={Pseudo-colouring an ECG enables lay people to detect QT-interval prolongation regardless of heart rate},
  author={Alahmadi, Alaa and Davies, Alan and Vigo, Markel and Jay, Caroline},
  journal={Plos one},
  volume={15},
  number={8},
  pages={e0237854},
  year={2020},
  publisher={Public Library of Science San Francisco, CA USA}
}

@article{snell2017prototypical,
  title={Prototypical networks for few-shot learning},
  author={Snell, Jake and Swersky, Kevin and Zemel, Richard},
  journal={Advances in neural information processing systems},
  volume={30},
  year={2017}
}

@article{EasyFSL,
   title={Easy Few-Shot Learning Library},
   author={Etienne, Bennequin},
    URL={https://github.com/sicara/easy-few-shot-learning},
    year={2024}
}

@article{gilchrist2009psychology,
  title={The psychology of vision},
  author={Gilchrist, James M},
  journal={Optometry: Science, Techniques and Clinical Management E-Book},
  pages={51},
  year={2009},
  publisher={Elsevier Health Sciences}
}

@article{bhuiyan2015t,
  title={The T-peak--T-end Interval as a marker of repolarization abnormality: A comparison with the QT interval for five different drugs},
  author={Bhuiyan, Tanveer A and Graff, Claus and Kanters, J{\o}rgen K and Nielsen, Jimmi and Melgaard, Jacob and Matz, J{\o}rgen and Toft, Egon and Struijk, Johannes J},
  journal={Clinical Drug Investigation},
  volume={35},
  pages={717--724},
  year={2015},
  publisher={Springer}
}

@article{morin2012relationships,
  title={Relationships between the T-peak to T-end interval, ventricular tachyarrhythmia, and death in left ventricular systolic dysfunction},
  author={Morin, Daniel P and Saad, Marc N and Shams, Omar F and Owen, J Sam and Xue, Joel Q and Abi-Samra, Freddy M and Khatib, Sammy and Nelson-Twakor, Onajefe S and Milani, Richard V},
  journal={Europace},
  volume={14},
  number={8},
  pages={1172--1179},
  year={2012},
  publisher={Oxford University Press}
}

@article{alahmadi2019can,
  title={Can laypeople identify a drug-induced QT interval prolongation? A psychophysical and eye-tracking experiment examining the ability of nonexperts to interpret an ECG},
  author={Alahmadi, Alaa and Davies, Alan and Vigo, Markel and Jay, Caroline},
  journal={Journal of the American Medical Informatics Association},
  volume={26},
  number={5},
  pages={404--411},
  year={2019},
  publisher={Oxford University Press}
}







\end{document}